%% file: main.tex
\documentclass{article} 
\usepackage{iclr2027_conference,times}

\input{math_commands.tex}

\usepackage{amsthm}

\theoremstyle{definition}

\theoremstyle{plain}

\usepackage{hyperref}
\usepackage{url}
\usepackage{graphicx}
\usepackage{booktabs}
\usepackage{algorithm}
\usepackage{algpseudocode}
\usepackage{multirow}
\usepackage{subcaption}
\title{CAG-WAM: Causal Action-Gain World Action Models for Robot Manipulation}

\author{Peizheng Guo\textsuperscript{1,2},
Cheng Zhang\textsuperscript{3},
Xingyu Zhang\textsuperscript{1,2},
Changwen Zheng\textsuperscript{1,2},
Wenwen Qiang\textsuperscript{1,2}\thanks{Corresponding Author}\\
\textsuperscript{1}National Key Laboratory of Space Integrated Information System, Institute of Software Chinese\\ Academy of Sciences, Beijing, China\\
\textsuperscript{2}University of Chinese Academy of Sciences, Beijing, China\\
\textsuperscript{3}Roban
}

\iclrfinalcopy 
\begin{document}

\maketitle
\lhead{Preprint}

\begin{abstract}
World Action Models (WAMs) combine action generation with future prediction, allowing predicted futures to support action evaluation and robot decision making. However, the future associated with a candidate action may contain action-induced and environment-driven changes. Directly evaluating the predicted future may not fully characterize the future change associated with the candidate action. For example, a candidate action that worsens a favorable environmental future may still yield a favorable outcome, while one that improves an unfavorable environmental future may still yield an unfavorable outcome. To analyze this issue, we formulate a Structural Causal Model (SCM) for action evaluation in WAMs. The SCM formalizes how action-induced and environment-driven changes jointly shape future evaluation, and reveals a backdoor path through which environmental evolution can also affect the future evaluation associated with a candidate action. Intervention and backdoor adjustment account for this path, providing a causal basis for characterizing the future change associated with each candidate action. Based on this analysis, we propose Causal Action-Gain World Action Model (CAG-WAM). For each candidate action, CAG-WAM predicts an action-conditioned future and an environmental future under the same current observation, and defines their evaluation difference as causal action gain (CAG) for characterizing the future change associated with the candidate action. During training, we incorporate CAG into Group Relative Policy Optimization (GRPO) to guide robot policy learning. During inference, we evaluate the initial action with CAG and, when needed, revise it using predicted future information. Experiments on multiple benchmarks support the effectiveness of CAG-WAM.
\end{abstract}

\section{Introduction}
\label{sec:intro}
World Action Models (WAMs) have recently been explored for robotic manipulation by combining action generation with future prediction~\citep{ye2026world,team2026motubrain,chen2026abot}. Given current visual and proprioceptive observations together with a task instruction, these models are trained to generate robot actions and predict future observations or scene evolution~\citep{liu2026oa,pan2026selfwam,zhu2026dswam}. Such future predictions provide information about the possible outcomes of generated actions and can inform subsequent action generation and decision making. Some recent WAMs further generate multiple candidate actions and predict the future associated with each candidate before execution~\citep{zhou2026tau_0,xue2026hitac}. These predicted futures can then be evaluated using task-related information, such as task progress or value estimates, to compare candidate actions and guide action selection~\citep{wang2026ea,lin2025storm}.

\begin{figure*}[t]
    \centering
    \includegraphics[width=\columnwidth]{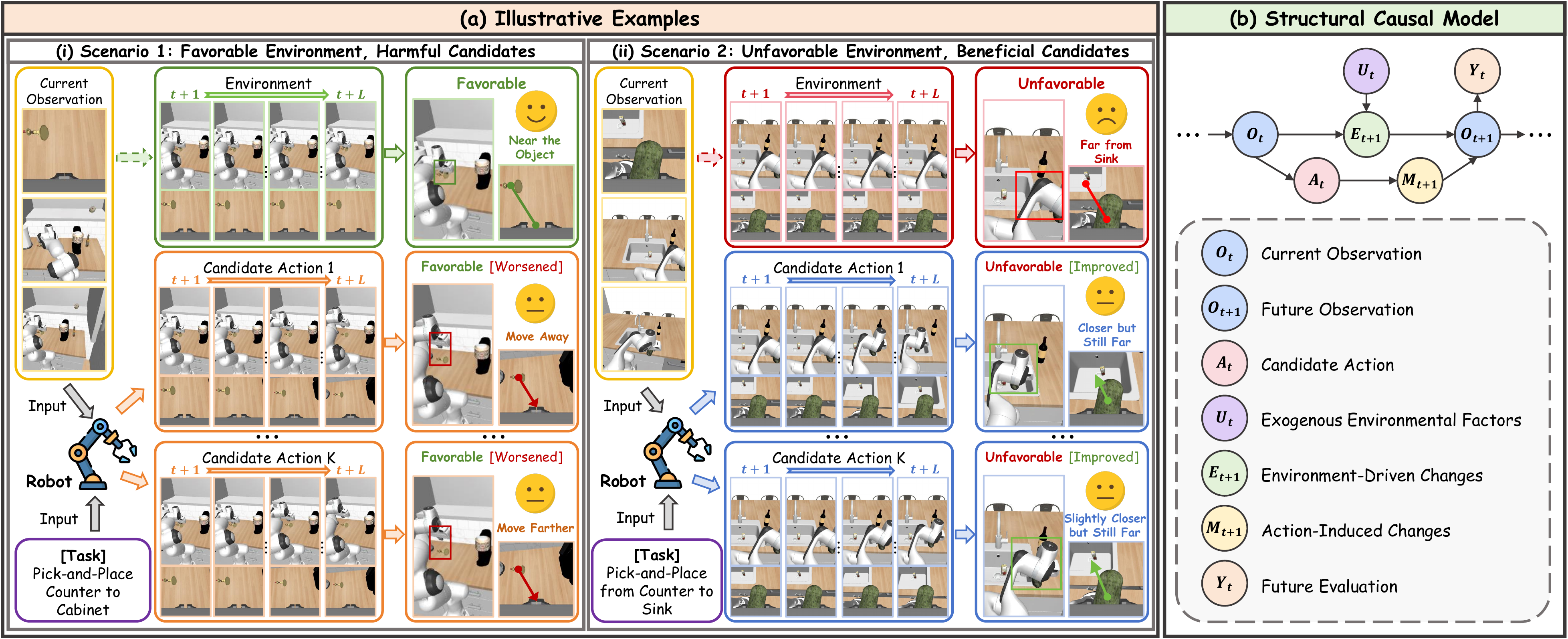}
    \caption{Motivation and causal analysis. (a) Illustrative cases where the overall future evaluation may not fully reflect the future change associated with a candidate action. (b) Structural causal model of action evaluation, including the backdoor path $\mathbf{A}_t\leftarrow\mathbf{O}_t\rightarrow\mathbf{E}_{t+1}\rightarrow\mathbf{O}_{t+1}\rightarrow Y_t$.}
    \label{fig:example+scm}
\end{figure*}

However, the predicted future associated with an action may contain changes from different sources. During robot--environment interaction, some future changes are associated with the action, while others may arise from environmental dynamics or external factors, such as passive object motion or external disturbances~\citep{liang2026exopredicator,jiang2026flowmo,gu2026geniworld,wang2026world}. Consequently, the overall future evaluation alone may not fully characterize the future change associated with the action.
For example, in \textbf{Scenario 1} of \textbf{Figure~\ref{fig:example+scm}a}, the environment evolves toward a favorable future in which the robot remains near the object. Candidate actions that move the robot away can still result in futures that appear favorable overall, despite worsening the state relative to the environmental evolution.
Conversely, in \textbf{Scenario 2}, the environment evolves toward an unfavorable future in which the robot remains far from the sink. Candidate actions that move the robot closer can still result in futures that appear unfavorable overall, despite improving the state relative to the environmental evolution.
A natural approach is to compare multiple candidates generated under the same observation~\citep{zhou2026tau_0,shao2024deepseekmath}.
Such comparison can distinguish relatively better and worse candidates within the current set, but it still evaluates each candidate relative to the other sampled actions.
For example, in \textbf{Scenario 1}, a relative comparison may prefer the candidate that moves the robot away over the one that moves it farther away, although both candidates worsen the favorable environmental future.
In \textbf{Scenario 2}, the candidate that moves the robot only slightly closer may be less preferred than the one that moves it farther toward the sink, although both candidates improve the unfavorable environmental future.
Thus, we focus on a central question: \textbf{how can a WAM characterize the future change associated with each candidate action when the predicted future contains changes from different sources, and use this information to guide policy learning and action generation?}

To address this problem, we analyze action evaluation from a causal perspective.
We construct the Structural Causal Model (SCM) shown in \textbf{Figure~\ref{fig:example+scm}b} to describe the relationships among the current observation, candidate action, different sources of future change, and future evaluation. Specifically, the current observation $\mathbf{O}_t$ influences both the generation of a candidate action $\mathbf{A}_t$ and the subsequent environment-driven changes $\mathbf{E}_{t+1}$. Meanwhile, the candidate action gives rise to action-induced changes $\mathbf{M}_{t+1}$, and these two sources of change jointly shape the future observation $\mathbf{O}_{t+1}$, which is then evaluated as $Y_t$.
In addition to its effect through the action-induced changes, $\mathbf{A}_t$ and $Y_t$ are also connected through the backdoor path $\mathbf{A}_t\leftarrow\mathbf{O}_t\rightarrow\mathbf{E}_{t+1}\rightarrow\mathbf{O}_{t+1}\rightarrow Y_t$~\citep{pearl2009causality,peters2017elements}.
This path reflects that the current observation affects both which action is generated and how the environment may evolve.
Therefore, the future evaluation associated with an action may reflect both action-induced and environment-driven changes, making it difficult to characterize the future change associated with the action from the overall future evaluation alone.
Backdoor adjustment provides a principled way to account for this backdoor path. Specifically, the intervention $do(\mathbf{A}_t=\mathbf{a})$ fixes the action to $\mathbf{a}$ and cuts the incoming edge $\mathbf{O}_t\rightarrow\mathbf{A}_t$ in the causal graph, thereby blocking this path~\citep{pearl2009causality,hernan2010causal}.
This analysis therefore provides a causal basis for characterizing the future change associated with each candidate action.

Motivated by this analysis, we propose Causal Action-Gain World Action Model (CAG-WAM), which characterizes the future change associated with each candidate action and uses this information to guide policy learning and action generation. CAG-WAM consists of a Future-Conditioned Action Model (FC-AM) and a Paired-Future World Model (PF-WM), with a shared future evaluator used to evaluate the predicted futures. Given the current observation and task instruction, the FC-AM generates candidate action chunks, while the PF-WM predicts an action-conditioned future for each candidate and an environmental future from the same current observation. The action-conditioned future describes the predicted outcome under the candidate action, while the environmental future describes how the scene may evolve without applying that action. The two futures are evaluated by the same future evaluator, and their evaluation difference defines the causal action gain (CAG), which characterizes the future change associated with the candidate action relative to environmental evolution.
During training, we use CAG as the advantage in group relative policy optimization (GRPO)~\citep{shao2024deepseekmath}, so that policy updates can reflect the future change associated with each candidate action.
During inference, the initial action is evaluated using the same paired-future comparison. When its CAG falls below the revision threshold, future information from the action-conditioned future and its difference from the environmental future is provided to the same action model for a second generation. The revised action is then combined with the initial action before execution.
In this way, CAG-WAM can emphasize the future change associated with each candidate action in both policy optimization during training and action revision during inference.
Extensive experiments support the effectiveness of CAG-WAM.

The main contributions of this paper are summarized as follows:
(\textbf{i}) We formulate action evaluation in WAMs from a causal perspective and construct a structural causal model that describes how action-induced and environment-driven changes jointly affect future evaluation and identifies the associated backdoor path.
(\textbf{ii}) We propose CAG-WAM, which introduces causal action gain to characterize the future change associated with each candidate action relative to environmental evolution, and uses this information to guide policy learning and action revision.
(\textbf{iii}) Experiments conducted on multiple benchmarks support the effectiveness of CAG-WAM.

\section{Related Work}
\label{sec:rw}
World Action Models (WAMs) integrate future prediction with robot action generation, allowing predicted outcomes to support manipulation decisions.
Motubrain~\citep{team2026motubrain} jointly models video dynamics and robot actions within a unified generative framework.
$\tau_0$-WM~\citep{zhou2026tau_0} evaluates candidate actions through action-conditioned future prediction and task-progress estimation, and uses predicted futures for action refinement.
SelfWAM~\citep{pan2026selfwam} strengthens the dependence of predicted futures on executed actions through explicit action conditioning and self-mask supervision.
HiTac-WAM~\citep{xue2026hitac} incorporates predicted tactile futures for candidate-action ranking in contact-rich manipulation.
These studies model action-conditioned futures and use predicted outcomes for candidate evaluation, action selection, or refinement.
Our work instead focuses on action-conditioned future evaluation from a causal perspective, considering that the predicted future may contain changes associated with both the candidate action and environmental evolution.
We therefore characterize the future change associated with each candidate action to guide policy learning and action generation.

Causal reasoning has been applied to robot learning through interventions, counterfactual reasoning, and deconfounding to study the relationships among observations, actions, and task outcomes.
CREST~\citep{lee2021causal} uses interventions in internal models to identify task-relevant state variables for robot manipulation policies.
RoCoDA~\citep{ameperosa2025rocoda} exploits causal invariance to construct counterfactual demonstrations by modifying task-irrelevant environmental factors.
CofactVLA~\citep{zhang2026cofactvla} models visual confounding in action generation by vision-language-action (VLA) models and introduces counterfactual intervention to reduce spurious visual dependencies.
FACT~\citep{peng2026fact} introduces causal training into WAMs by conditioning future video and task-progress prediction on executed actions and incorporating failure rollouts to model action-dependent consequences.
These studies apply causal reasoning to different aspects of robot learning, including state-variable identification, counterfactual data construction, action generation, and future prediction.
Our work focuses on future evaluation in WAMs and constructs a structural causal model to analyze how candidate actions and environmental evolution jointly influence future evaluation.
Based on this analysis, we characterize the future change associated with each candidate action, rather than relying on the overall future evaluation alone, and use this information to guide policy learning and action generation.

\section{Problem Settings and Analysis}
\label{sec:problem_analysis}

\paragraph{Problem Setting}
\label{sec:problem_setting}
We consider a World Action Model (WAM) for robotic manipulation that supports both action generation and future prediction~\citep{ye2026world,team2026motubrain,zhou2026tau_0}.

At time step $t$, let $\mathbf{O}_t$ denote the observations available to the robot, including visual and proprioceptive observations.
Given the current observation $\mathbf{O}_t$ and task instruction $\mathbf{c}$, the action policy samples $K$ candidate actions $\mathbf{A}_t^{i}\sim\pi_{\theta}(\cdot | \mathbf{O}_t,\mathbf{c})$, where $i=1,\ldots,K$ and $\theta$ denotes the policy parameters.
For each candidate action, the world model predicts its corresponding future as
$\hat{\mathbf{O}}_{t+1:t+H}^{\,i}\sim p_{\phi}(\cdot |\mathbf{O}_t,\mathbf{A}_t^{i},\mathbf{c})$,
where $\hat{\mathbf{O}}_{t+1:t+H}^{\,i}$ denotes the predicted future observation sequence over a prediction horizon $H$, and $\phi$ denotes the world-model parameters.

We evaluate each candidate's predicted future as $Y_t^{i}=\mathcal{V}(\hat{\mathbf{O}}_{t+1:t+H}^{\,i},\mathbf{c})$, where $\mathcal{V}(\cdot)$ denotes a generic future evaluator that returns a scalar task-related score, such as a task reward, value estimate, task-progress score, or a combination thereof.
The resulting future evaluations can then be used for candidate comparison, action selection, or subsequent policy optimization.

\paragraph{Motivation Analysis}
\label{sec:motivation}
WAMs can use the future evaluations $Y_t^{i}$ defined above for candidate comparison and action selection~\citep{zhou2026tau_0,xue2026hitac,wang2026ea,lin2025storm}.
However, the future predicted for a candidate action may reflect changes from multiple sources.
During robot--environment interaction, some changes may be associated with the robot action, while others may arise from environmental dynamics or external factors, such as passive object motion or external disturbances~\citep{liang2026exopredicator,jiang2026flowmo,gu2026geniworld,wang2026world}.
Consequently, the evaluation of the resulting future may not by itself characterize the change associated with the candidate action.
As illustrated in \textbf{Figure~\ref{fig:example+scm}a}, \textbf{Scenario 1} considers a favorable environment in which the robot would remain near the target object without applying the candidate action.
A candidate action that moves the robot away may still lead to a future that appears favorable overall, although it worsens the state relative to the environmental future.
In contrast, \textbf{Scenario 2} considers an unfavorable environment in which the robot would remain far from the sink without applying the candidate action.
A candidate action that moves the robot closer may still lead to a future that appears unfavorable overall, although it improves the state relative to the environmental future.
These cases show that the overall evaluation of the predicted future and the change associated with the candidate action may not be equivalent.

A natural approach is to compare multiple candidates generated from the same current observation~\citep{zhou2026tau_0,shao2024deepseekmath}.
Such comparisons rank the sampled candidates according to their future evaluations: $Y_t^{i}>Y_t^{j}$ favors candidate $i$ over candidate $j$.
\textbf{Figure~\ref{fig:example+scm}a} illustrates the limitation of relying on this relative ranking alone.
In \textbf{Scenario 1}, the candidate that moves the robot away can be preferred over the candidate that moves it farther away, although both actions worsen the favorable environmental future.
Conversely, in \textbf{Scenario 2}, the candidate that moves the robot only slightly closer can be less preferred than the candidate that moves it farther toward the sink, although both actions improve the unfavorable environmental future.
Thus, candidate-relative comparison indicates which sampled actions are relatively preferred, but does not by itself characterize whether each individual action improves or worsens the future relative to environmental evolution.
Therefore, we focus on a central question: \textbf{how can a WAM characterize the future change associated with each candidate action when the predicted future contains changes from multiple sources, and use this information to guide policy learning and action generation?}

\begin{figure*}[t]
    \centering
    \includegraphics[width=\columnwidth]{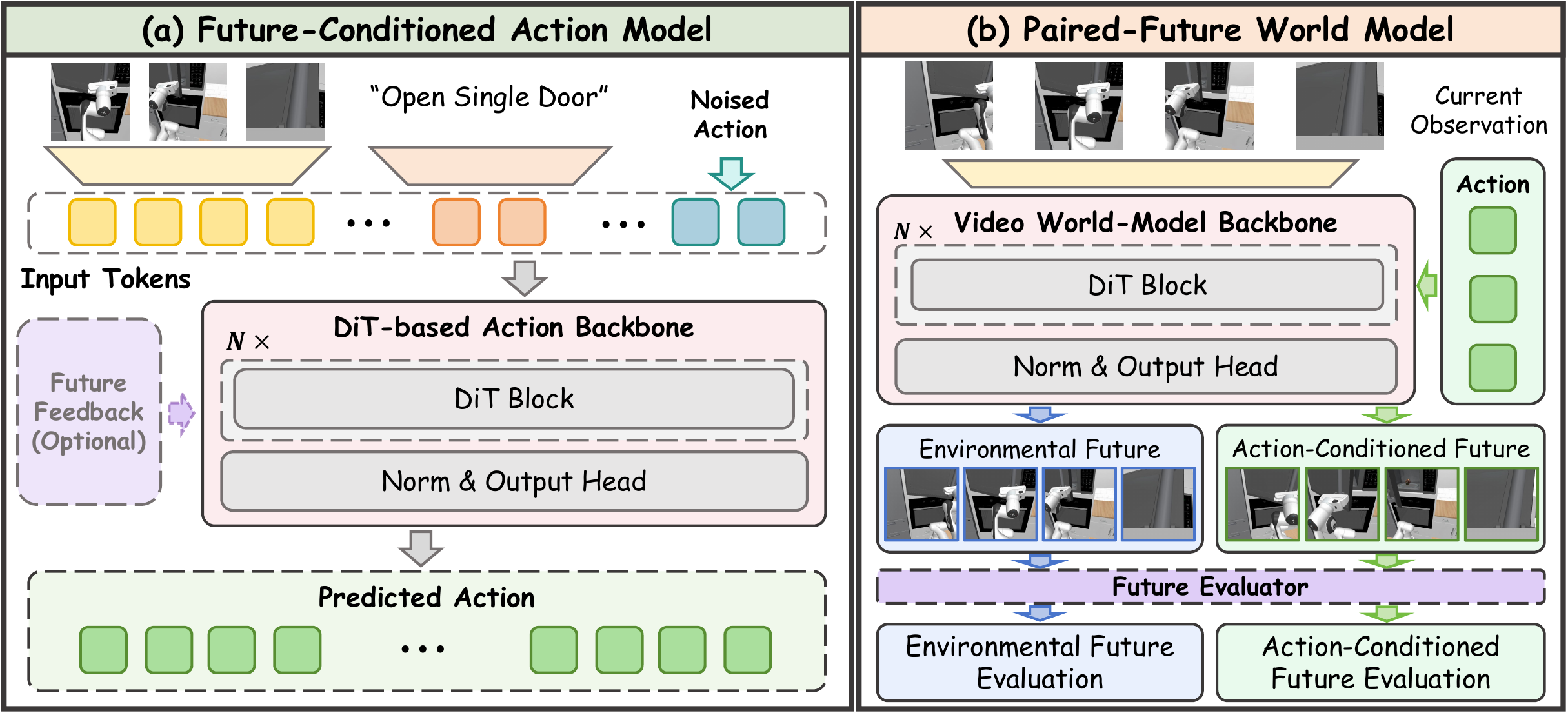}
    \caption{CAG-WAM Architecture. (a) Future-Conditioned Action Model generates action chunks from the current observation and task instruction, and can additionally incorporate future feedback for action revision. (b) Paired-Future World Model predicts action-conditioned and environmental futures under different action conditions and evaluates them using the same future evaluator.}
    \label{fig:architecture}
\end{figure*}

\paragraph{Causal Analysis}
\label{sec:causal_analysis}
To address this issue, we analyze action evaluation in WAMs from a causal perspective.
We construct a Structural Causal Model (SCM) to describe the relationships among the current observation, candidate action, different sources of future change, and future evaluation.

As shown in \textbf{Figure~\ref{fig:example+scm}b}, let $\mathbf{O}_t$ denote the current observation, $\mathbf{A}_t$ a candidate action, and $\mathbf{U}_t$ exogenous environmental factors that are not directly controlled by the current action.
We use $\mathbf{E}_{t+1}$ to denote environment-driven changes and $\mathbf{M}_{t+1}$ to denote action-induced changes.
The environment-driven changes $\mathbf{E}_{t+1}$ depend on the current observation $\mathbf{O}_t$ and exogenous environmental factors $\mathbf{U}_t$, while the candidate action $\mathbf{A}_t$ gives rise to the action-induced changes $\mathbf{M}_{t+1}$.
These two sources of change jointly shape the future observation $\mathbf{O}_{t+1}$, which is subsequently evaluated as $Y_t$ under the task instruction $\mathbf{c}$.
Thus, the predicted future may contain changes arising from both the candidate action and environmental evolution.

Because the candidate action $\mathbf{A}_t$ is generated conditioned on the current observation $\mathbf{O}_t$, $\mathbf{O}_t$ influences which action is generated and also affects the future through the environment-driven changes $\mathbf{E}_{t+1}$.
The candidate action $\mathbf{A}_t$ affects $Y_t$ through the action-induced changes $\mathbf{M}_{t+1}$, while $\mathbf{A}_t$ and $Y_t$ are also connected through the backdoor path $\mathbf{A}_t\leftarrow\mathbf{O}_t\rightarrow\mathbf{E}_{t+1}\rightarrow\mathbf{O}_{t+1}\rightarrow Y_t$.
This path reflects that the same current observation influences both which action is generated and how the environment may evolve.
Consequently, the future evaluation associated with a candidate action may reflect both action-induced and environment-driven changes, making it difficult to characterize the future change associated with the action from the overall future evaluation alone.
For clarity, we condition on the task instruction $\mathbf{c}$ throughout this analysis and omit it from the causal notation.
Formally, the expected future evaluation associated with action $\mathbf{a}$ can be written as
\begin{equation}
\mathbb{E}[Y_t\mid\mathbf{A}_t=\mathbf{a}]
=
\int
\mathbb{E}[Y_t\mid\mathbf{A}_t=\mathbf{a},\mathbf{O}_t=\mathbf{o}]
p(\mathbf{o}\mid\mathbf{A}_t=\mathbf{a})d\mathbf{o},
\label{eq:observational_evaluation}
\end{equation}
where $\mathbf{a}$ and $\mathbf{o}$ denote possible realizations of $\mathbf{A}_t$ and $\mathbf{O}_t$, respectively.
Because candidate actions are generated from $\mathbf{O}_t$, $p(\mathbf{o}|\mathbf{A}_t=\mathbf{a})$ can vary with the action.
Since $\mathbf{O}_t$ also affects $Y_t$ through $\mathbf{E}_{t+1}$, this evaluation may also reflect the influence of the backdoor path.

Backdoor adjustment provides a principled way to account for this path through intervention on the action variable~\citep{pearl2009causality,hernan2010causal}.
Specifically, the intervention $do(\mathbf{A}_t=\mathbf{a})$ fixes the action to $\mathbf{a}$ and cuts the incoming edge $\mathbf{O}_t\rightarrow\mathbf{A}_t$ in the causal graph, thereby blocking the backdoor path.
Under the SCM, the corresponding interventional evaluation can be expressed as
\begin{equation}
\mathbb{E}[Y_t\mid do(\mathbf{A}_t=\mathbf{a})]
=
\int
\mathbb{E}[Y_t\mid\mathbf{A}_t=\mathbf{a},\mathbf{O}_t=\mathbf{o}]
p(\mathbf{o})d\mathbf{o}.
\label{eq:backdoor_adjustment}
\end{equation}
Compared with \textbf{Eq.~\ref{eq:observational_evaluation}}, the action-dependent observation distribution $p(\mathbf{o}|\mathbf{A}_t=\mathbf{a})$ is replaced by the observation distribution $p(\mathbf{o})$, thereby reducing the influence associated with the backdoor path and allowing
the action evaluation to better reflect the future change associated with the action.

Overall, this causal analysis shows that action-induced and environment-driven changes jointly shape future evaluation, while the SCM reveals a backdoor path connecting the candidate action and future evaluation through the current observation and environment-driven changes.
By accounting for this path through intervention and backdoor adjustment, this analysis provides a causal basis for characterizing the future change associated with each candidate action.

\section{Methodology}
\label{sec:method}
In this section, we propose Causal Action-Gain World Action Model (CAG-WAM) to characterize the future change associated with each candidate action and use this information to guide both policy learning and action generation.
\textbf{Section~\ref{sec:architecture}} introduces the overall architecture of CAG-WAM and the interaction among action generation, future prediction, and future evaluation.
\textbf{Section \ref{sec:train}} presents how causal action gain is computed and incorporated into policy optimization.
\textbf{Section \ref{sec:inference}} describes how the predicted future information is further used to revise actions during inference.

\begin{figure*}[t]
    \centering
    \includegraphics[width=\columnwidth]{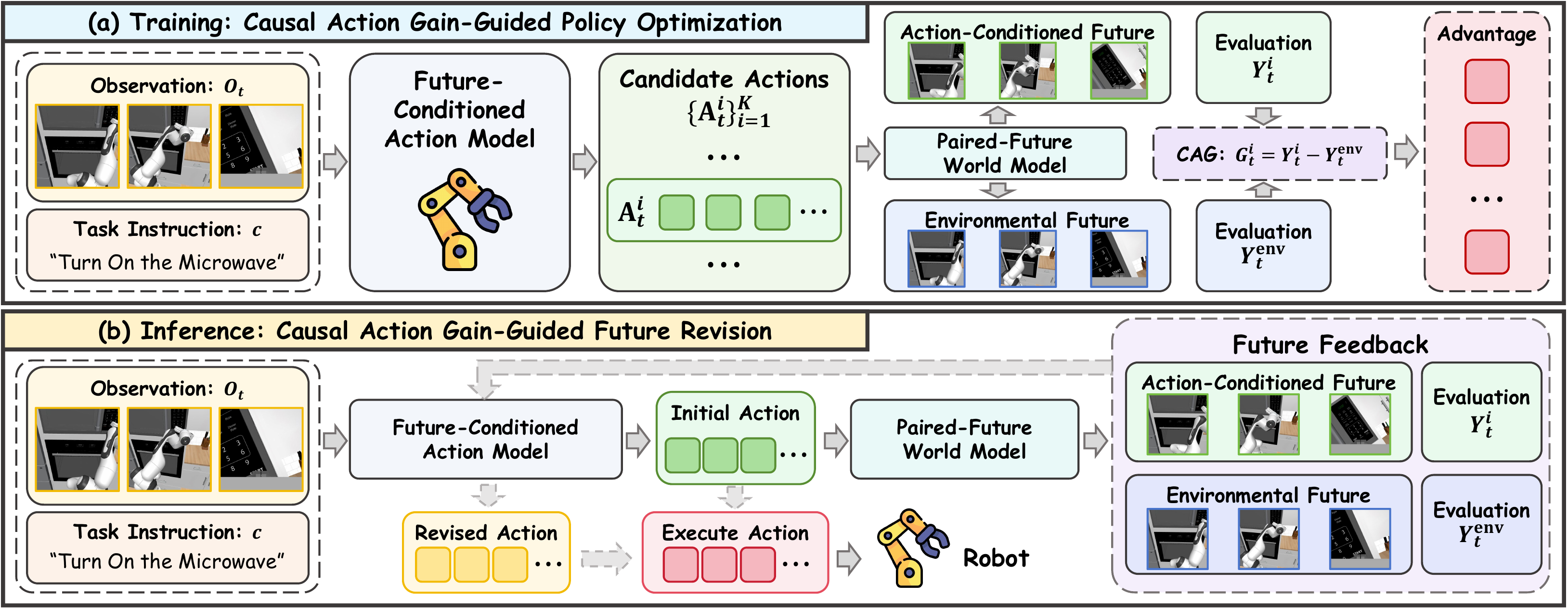}
    \caption{Training and inference framework of CAG-WAM. (a) Training: candidate actions are evaluated by comparing their action-conditioned futures with the environmental future, and the resulting causal action gain is used as the policy-optimization advantage. (b) Inference: the initial action is evaluated using the same paired-future comparison; when its causal action gain falls below the revision threshold, future feedback is used for an additional action generation before execution.}
    \label{fig:framework}
\end{figure*}

\subsection{Architecture: Causal Action-Gain World Action Model}
\label{sec:architecture}
As illustrated in \textbf{Figure~\ref{fig:architecture}}, CAG-WAM consists of a Future-Conditioned Action Model (FC-AM) and a Paired-Future World Model (PF-WM).
(i) The FC-AM adopts a flow-matching formulation with a DiT backbone. Given the current observation $\mathbf{O}_t$ and task instruction $\mathbf{c}$, it generates $K$ candidate action chunks $\{\mathbf{A}_t^i\}_{i=1}^{K}$, where $\mathbf{A}_t^i=[\mathbf{a}_t^i,\ldots,\mathbf{a}_{t+L-1}^i]$ and $L$ denotes the action-chunk length.
Besides generating actions from the current observation, the same FC-AM can also incorporate predicted future information as an additional condition for subsequent action generation.
(ii) The PF-WM is an action-conditioned video world model that predicts future observations under different action conditions.
For each candidate action $\mathbf{A}_t^i$, it predicts the corresponding action-conditioned future $\hat{\mathbf{O}}_{t+1:t+H}^{\,i}$, where $H$ denotes the prediction horizon.
From the same current observation, the PF-WM also predicts an environmental future $\hat{\mathbf{O}}_{t+1:t+H}^{\mathrm{env}}$, describing how the scene may evolve without applying the candidate action.
Each action-conditioned future and the environmental future are generated from the same current observation and task instruction, while differing in their action conditions.
They are evaluated by the same future evaluator, yielding $Y_t^i=\mathcal{V}(\hat{\mathbf{O}}_{t+1:t+H}^{\,i},\mathbf{c})$ and $Y_t^{\mathrm{env}}=\mathcal{V}(\hat{\mathbf{O}}_{t+1:t+H}^{\,\mathrm{env}},\mathbf{c})$.
Their evaluation difference is used to guide policy optimization during training, while information from the predicted futures is used to condition the FC-AM for action revision during inference.

\subsection{Training: Causal Action Gain-Guided Policy Optimization}
\label{sec:train}
In this section, we introduce how the action policy is trained using causal action gain.
We first compute the causal action gain of each candidate by comparing its predicted future with the environmental future.
We then incorporate the causal action gain into GRPO, allowing the future change associated with each candidate action to guide policy learning.

\subsubsection{Causal Action Gain}
\label{sec:cag_estimate}
Given the $K$ candidate action chunks $\{\mathbf{A}_t^i\}_{i=1}^{K}$ generated from the current observation $\mathbf{O}_t$, we characterize the future change associated with each candidate through paired future prediction. 
For each candidate action $\mathbf{A}_t^i$, the PF-WM predicts its action-conditioned future $\hat{\mathbf{O}}_{t+1:t+H}^{\,i}$, which describes the predicted outcome under the candidate action.
From the same current observation, the PF-WM also predicts an environmental future $\hat{\mathbf{O}}_{t+1:t+H}^{\,\mathrm{env}}$, which describes how the scene may evolve without applying the candidate action. 
The action-conditioned future and environmental future are generated from the same current observation and task instruction, providing a matched comparison for evaluating the change associated with the candidate action.

We evaluate both futures using the same future evaluator $\mathcal{V}$: $Y_t^i=\mathcal{V}(\hat{\mathbf{O}}_{t+1:t+H}^{\,i},\mathbf{c})$ and $Y_t^{\mathrm{env}}= \mathcal{V}(\hat{\mathbf{O}}_{t+1:t+H}^{\,\mathrm{env}},\mathbf{c})$. Here, $Y_t^i$ and $Y_t^{\mathrm{env}}$ denote the evaluations of the action-conditioned future and environmental future, respectively. Following \textbf{Section~\ref{sec:causal_analysis}}, we define the causal action gain of candidate action $\mathbf{A}_t^i$ as $G_t^i=Y_t^i-Y_t^{\mathrm{env}}$.
A positive $G_t^i$ means that the action-conditioned future receives a higher evaluation than the environmental future, while a negative value means that it receives a lower evaluation. 
The magnitude of $G_t^i$ reflects the evaluation difference between the two futures.
Thus, CAG uses the paired future evaluations to characterize the future change associated with the candidate action relative to environmental evolution (see the \textbf{Appendix} for further discussion).

\subsubsection{Causal Action Gain-Guided Optimization}
\label{sec:cag_optimization}
After obtaining the causal action gain, we use it to optimize the action policy.
Group Relative Policy Optimization (GRPO)~\citep{shao2024deepseekmath} provides an efficient way to optimize policies from grouped samples without requiring an additional critic model, and has recently been extended to embodied reasoning and robot policy learning~\citep{liu2026flow,zhao2025embodied,cao2026z}.
We incorporate the causal action gain into GRPO so that the future change associated with each candidate action can directly inform policy optimization.

Specifically, given the current observation $\mathbf{O}_t$ and task instruction $\mathbf{c}$, the flow-based action policy samples $K$ candidate action chunks $\{\mathbf{A}_t^i\}_{i=1}^{K}$.
For each candidate action chunk $\mathbf{A}_t^i$, the paired future evaluation provides a causal action gain $G_t^i$.
We use this gain as the advantage of the corresponding candidate action, i.e., $\Delta_{t,\mathrm{CAG}}^i=G_t^i$.

We then optimize the action policy using the clipped GRPO objective:
\begin{equation}
\resizebox{0.94\linewidth}{!}{$
\displaystyle
\mathcal{J}_{\mathrm{CAG}}(\theta)=\mathbb{E}\left[\frac{1}{K}\sum_{i=1}^{K}\frac{1}{T}\sum_{\tau=1}^{T}(\min(\rho_{t,\tau}^{i}(\theta)G_t^i,\operatorname{clip}(\rho_{t,\tau}^{i}(\theta),1-\epsilon,1+\epsilon)G_t^i)-\beta D_{\mathrm{KL}}(\pi_{\theta}\Vert\pi_{\mathrm{ref}}))\right]
$}
\label{eq:cag_grpo_objective}
\end{equation}
where $T$ denotes the number of flow-generation steps, $\epsilon$ is the clipping coefficient, $\beta$ controls KL regularization, and $\pi_{\mathrm{ref}}$ denotes the reference policy.
Let $\mathbf{X}_{t,\tau}^{i}$ denote the intermediate action variable of candidate $i$ at generation step $\tau$.
The importance ratio between the current policy and the sampling policy is computed as $\rho_{t,\tau}^{i}(\theta)=\frac{p_{\theta}(\mathbf{X}_{t,\tau-1}^{i}|\mathbf{X}_{t,\tau}^{i},\mathbf{O}_t,\mathbf{c})}{p_{\theta_{\mathrm{old}}}(\mathbf{X}_{t,\tau-1}^{i}|\mathbf{X}_{t,\tau}^{i},\mathbf{O}_t,\mathbf{c})}$ following the probabilistic formulation for flow-based policy optimization~\citep{liu2026flow}  (see the \textbf{Appendix} for more details).

In this way, candidate action chunks with positive causal action gains tend to be encouraged, whereas those with negative gains tend to be suppressed. The future change associated with each candidate action, as characterized by the causal action gain, can guide the optimization of the action policy.

\subsection{Inference: Causal Action Gain-Guided Future Revision}
\label{sec:inference}
During inference, we use predicted future information to revise an initial action before execution when its causal action gain falls below a revision threshold.
For such actions, information about the difference between the action-conditioned future and the environmental future is used for a second action generation.
The revised action is then combined with the initial action before execution.

Specifically, given the current observation $\mathbf{O}_t$ and task instruction $\mathbf{c}$, the policy first generates an initial action chunk $\mathbf{A}_t^{\mathrm{init}}$.
The PF-WM then predicts its action-conditioned future $\hat{\mathbf{O}}_{t+1:t+H}^{\mathrm{init}}$ and the environmental future $\hat{\mathbf{O}}_{t+1:t+H}^{\mathrm{env}}$.
Using the same future evaluator as in training, we obtain $Y_t^{\mathrm{init}}$ and $Y_t^{\mathrm{env}}$ and compute the causal action gain $G_t^{\mathrm{init}}=Y_t^{\mathrm{init}}-Y_t^{\mathrm{env}}$.
When $G_t^{\mathrm{init}}<\delta$, we perform an additional future-conditioned action generation, where $\delta$ denotes the revision threshold.

For actions that require revision, we use the internal representations of the two predicted futures to construct the feedback for the second generation.
Let $\hat{\mathbf{Z}}_{t+1:t+H}^{\mathrm{init}}$ and $\hat{\mathbf{Z}}_{t+1:t+H}^{\mathrm{env}}$ denote the internal representations of the action-conditioned future and the environmental future, respectively.
The former represents the predicted future under the initial action, while their difference captures the discrepancy between the action-conditioned and environmental futures.
We construct the future feedback as
$\hat{\mathbf{Z}}_{t+1:t+H}^{\mathrm{fb}}=\hat{\mathbf{Z}}_{t+1:t+H}^{\mathrm{init}}+\gamma(\hat{\mathbf{Z}}_{t+1:t+H}^{\mathrm{init}}-\hat{\mathbf{Z}}_{t+1:t+H}^{\mathrm{env}})$, where $\gamma$ controls the contribution of the difference between the two future representations.
The resulting future feedback is provided to the FC-AM together with the current observation and task instruction.
The FC-AM then performs another complete generation to obtain a revised action chunk $\mathbf{A}_t^{\mathrm{rev}}$.

We then combine the revised action with the initial action before execution:
\begin{equation}
    \mathbf{A}_t^{\mathrm{exec}}=\mathbf{A}_t^{\mathrm{init}}+\lambda(\mathbf{A}_t^{\mathrm{rev}}-\mathbf{A}_t^{\mathrm{init}}),\quad \lambda\in[0,1],
    \label{eq:action_fusion}
\end{equation}
where $\lambda$ controls the contribution of the revised action to the final output.

Through this inference procedure, predicted future information is selectively incorporated into action generation according to the causal action gain. For actions that require revision, the difference between the action-conditioned and environmental future representations provides additional information for generating a revised action, while the final fusion controls how strongly the revised action modifies the initial policy output.

\begin{table*}[t]
\centering
\caption{
Success rates (\%) on the 24-task RoboCasa benchmark. The tasks are grouped into seven manipulation skill families: Pick\&Place (8 tasks), Door (4 tasks), Drawer (2 tasks), Knob (2 tasks), Lever (3 tasks), Button (3 tasks), and Insertion (2 tasks). Avg. denotes the average success rate over all 24 tasks. Full results are provided in the \textbf{Appendix}.}
\label{tab:robocasa_main}
\resizebox{\textwidth}{!}{
\begin{tabular}{lcccccccc}
\toprule
\textbf{Method} & \textbf{Pick\&Place} & \textbf{Door} & \textbf{Drawer} & \textbf{Knob} & \textbf{Lever} & \textbf{Button} & \textbf{Insertion} & \textbf{Avg.} \\
\midrule
3D Diffuser Actor \citep{ke20243d} & 0.0 & 3.5 & 0.0 & 6.0 & 20.0 & 15.3 & 0.0 & 5.5 \\
DP3 \citep{ze20243d} & 1.5 & 40.5 & 42.0 & 15.0 & 42.7 & 36.0 & 11.0 & 22.8 \\
Diffusion Policy (ResNet) \citep{chi2025diffusion} & 6.2 & 67.0 & 73.0 & 27.0 & 65.3 & 68.7 & 27.0 & 40.6 \\
Diffusion Policy (CLIP) \citep{chi2025diffusion} & 6.0 & 61.5 & 78.0 & 18.0 & 54.0 & 85.3 & 36.0 & 40.7 \\
GR00T N1 \citep{bjorck2025gr00t} & 22.6 & 58.0 & 89.0 & 41.5 & 63.0 & 78.0 & 48.0 & 49.7 \\
FPV \citep{donat2025towards} & 14.2 & 82.0 & 83.0 & 43.0 & 66.7 & 84.7 & 32.0 & 50.5 \\
\midrule
UVA \citep{li2025unifiedvideoactionmodel} & 20.8 & 77.0 & 50.0 & 32.0 & 63.3 & 91.3 & 49.0 & 50.0 \\
Video Policy (50 demos) \citep{liang2025video} & 50.5 & 90.5 & 71.0 & 18.0 & 67.3 & 91.3 & 49.0 & 63.2 \\
Video Policy (300 demos) \citep{liang2025video} & 47.2 & 86.0 & 77.0 & 27.0 & 75.3 & \textbf{97.3} & \textbf{64.0} & 65.7 \\
Cosmos Policy-Predict2.5 \citep{kim2026cosmos} & 55.8 & 89.5 & 96.0 & 49.0 & 84.0 & 91.3 & 48.0 & 71.5 \\
\midrule
\textbf{CAG-WAM (Ours)} & \textbf{60.8} & \textbf{92.0} & \textbf{96.0} & \textbf{52.0} & \textbf{84.0} & 92.0 & 51.0 & \textbf{74.2} \\
\bottomrule
\end{tabular}
}
\end{table*}

\begin{table*}[t]
\centering
\caption{Success rates (\%) on the four LIBERO task suites.}
\label{tab:libero_main}
\resizebox{0.8\textwidth}{!}{
\begin{tabular}{lccccc}
\toprule
\textbf{Method} & \textbf{Spatial} & \textbf{Object} & \textbf{Goal} & \textbf{Long} & \textbf{Avg.} \\
\midrule
Diffusion Policy~\citep{chi2025diffusion} & 78.3 & 92.5 & 68.3 & 50.5 & 72.4 \\
Octo~\citep{team2024octo} & 78.9 & 85.7 & 84.6 & 51.1 & 75.1 \\
OpenVLA~\citep{kim2024openvla} & 84.7 & 88.4 & 79.2 & 53.7 & 76.5 \\
TraceVLA~\citep{zheng2025tracevla} & 84.6 & 85.2 & 75.1 & 54.1 & 74.8 \\
SpatialVLA~\citep{qu2025spatialvla} & 88.2 & 89.9 & 78.6 & 55.5 & 78.1 \\
UniVLA~\citep{bu2025univla} & 96.5 & 96.8 & 95.6 & 92.0 & 95.2 \\
\midrule
LingBot-VA~\citep{li2026causal} & 98.5 & 99.6 & 97.2 & 98.5 & 98.5 \\
Cosmos Policy-Predict2~\citep{kim2026cosmos} & 98.1 & \textbf{100.0} & 98.2 & 97.6 & 98.5 \\
Cosmos Policy-Predict2.5~\citep{kim2026cosmos} & 98.7 & 99.7 & 98.3 & 96.6 & 98.3 \\
\midrule
\textbf{CAG-WAM (Ours)} & \textbf{99.5} & \textbf{100.0} & \textbf{100.0} & \textbf{99.6} & \textbf{99.8} \\
\bottomrule
\end{tabular}}
\end{table*}

\begin{figure*}[t]
    \centering
    \includegraphics[width=\columnwidth]{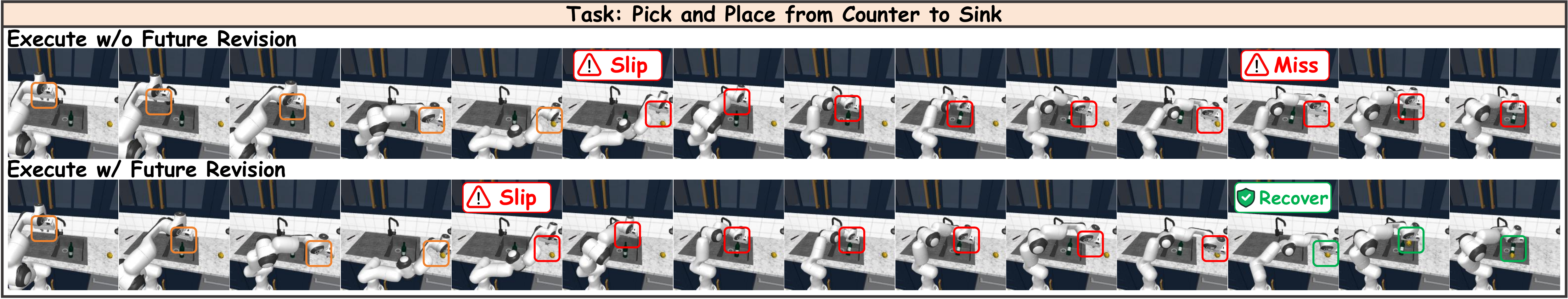}
    \caption{Visualization of CAG-WAM.}
    \label{fig:case}
\end{figure*}

\section{Experiments}
\label{sec:exp}
In this section, we conduct experiments to evaluate the effectiveness of CAG-WAM. More details, experiments, and analysis are provided in the \textbf{Appendix} due to space limitations.

\subsection{Experimental Setup}
\label{sec:exp_setup}

\paragraph{Benchmarks and Baselines}
We evaluate CAG-WAM on RoboCasa~\citep{nasiriany2024robocasa} and LIBERO~\citep{liu2023libero}, using task success rate as the primary metric. The RoboCasa evaluation covers seven manipulation skill families, while LIBERO contains four suites covering spatial, object, goal, and long-horizon settings. We compare CAG-WAM with many representative methods, including 3D Diffuser Actor~\citep{ke20243d}, DP3~\citep{ze20243d}, Diffusion Policy~\citep{chi2025diffusion}, GR00T N1~\citep{bjorck2025gr00t}, FPV~\citep{donat2025towards}, Octo~\citep{team2024octo}, OpenVLA~\citep{kim2024openvla}, TraceVLA~\citep{zheng2025tracevla}, SpatialVLA~\citep{qu2025spatialvla}, UniVLA~\citep{bu2025univla}, UVA~\citep{li2025unifiedvideoactionmodel}, Video Policy~\citep{liang2025video}, LingBot-VA~\citep{li2026causal}, Cosmos Policy-Predict2~\citep{kim2026cosmos}, and Cosmos Policy-Predict2.5~\citep{kim2026cosmos}.

\paragraph{Implementation Details}
We implement CAG-WAM based on publicly available robot policy and world-model codebases~\citep{ali2025world,kim2026cosmos}. 
During training, we use $K=8$ candidate actions with learning rates of $5\times10^{-5}$ for RoboCasa and $3\times10^{-6}$ for LIBERO.
During inference, we set $\delta=0.05$, $\gamma=1.0$, and $\lambda=0.30$ for threshold, future feedback, and action fusion, respectively.
More details are provided in the \textbf{Appendix}.

\subsection{Main Results}
\label{sec:exp_result}
\paragraph{Performance on RoboCasa}
\textbf{Table~\ref{tab:robocasa_main}} reports the results on RoboCasa.
CAG-WAM achieves an average success rate of $74.2\%$ across the 24 RoboCasa tasks.
These tasks span seven manipulation skill families: Pick\&Place, Door, Drawer, Knob, Lever, Button, and Insertion, covering diverse manipulation settings.
The task-level results in the \textbf{Appendix} further show that the performance improvements are distributed across multiple individual tasks and skill categories.
These results demonstrate the effectiveness of CAG-WAM across diverse RoboCasa manipulation settings.

\paragraph{Performance on Libero}
\textbf{Table~\ref{tab:libero_main}} reports the results on the four LIBERO suites.
CAG-WAM achieves an average success rate of $99.8\%$, with strong performance across all four task suites.
It reaches $100\%$ success on both Object and Goal, while maintaining high performance on Spatial and Long.
These results support the effectiveness of CAG-WAM across manipulation settings involving spatial relations, object variations, task goals, and long-horizon execution.

\subsection{Analysis}
\label{sec:exp_analysis}

\paragraph{Visualization and Case Study}
We present a manipulation example in \textbf{Figure~\ref{fig:case}}. Without future revision, an intermediate slip leads to an inaccurate grasp and eventually causes the action execution to miss the target. With future revision, the policy can adjust the subsequent action after the slip and recover to complete the task.
More results are provided in the \textbf{Appendix}.

\section{Conclusion}
\label{sec:conclusion}

In this work, we study action evaluation in World Action Models from a causal perspective.
Our analysis considers how action-induced and environment-driven changes can jointly shape future evaluation, such that the overall future evaluation may not fully characterize the future change associated with a candidate action.
To address this issue, we construct a structural causal model and propose CAG-WAM, which compares paired action-conditioned and environmental futures and defines their evaluation difference as causal action gain for the future change associated with the candidate action.
During training, CAG is incorporated into GRPO to guide robot policy learning. During inference, CAG-WAM evaluates the initial action with CAG and, when needed, revises it using predicted future information.
Extensive experiments support the effectiveness of CAG-WAM.

\section*{AI use statement}
In this work, we used generative AI tools for grammar and language polishing. They were not involved in research ideation, methodology, analysis, or substantive writing. The authors take full responsibility for all contents of the paper.

\bibliography{iclr2027_conference}
\bibliographystyle{iclr2027_conference}

\end{document}

%% file: math_commands.tex
\usepackage{amsmath,amsfonts,bm}

\def\eqref#1{equation~\ref{#1}}

\def\1{\bm{1}}

\DeclareMathAlphabet{\mathsfit}{\encodingdefault}{\sfdefault}{m}{sl}
\SetMathAlphabet{\mathsfit}{bold}{\encodingdefault}{\sfdefault}{bx}{n}



%% file: iclr2027_conference.bib
@article{ye2026world,
  title={World action models are zero-shot policies},
  author={Ye, Seonghyeon and Ge, Yunhao and Zheng, Kaiyuan and Gao, Shenyuan and Yu, Sihyun and Kurian, George and Indupuru, Suneel and Tan, You Liang and Zhu, Chuning and Xiang, Jiannan and others},
  journal={arXiv preprint arXiv:2602.15922},
  year={2026}
}

@article{team2026motubrain,
  title={Motubrain: An advanced world action model for robot control},
  author={Team, MotuBrain and Xiang, Chendong and Bao, Fan and Liu, Haitian and Tan, Hengkai and Bi, Hongzhe and Li, James and Liu, Jiabao and Pang, Jingrui and Jing, Kiro and others},
  journal={arXiv preprint arXiv:2604.27792},
  year={2026}
}

@article{liu2026oa,
  title={Oa-wam: Object-addressable world action model for robust robot manipulation},
  author={Liu, Yushan and Sun, Peibo and Li, Shoujie and Xie, Yifan and Zhang, Lingfeng and Chao, Xintao and Dong, Shiyuan and Chen, Fang and Zhang, Xiao-Ping and Ding, Wenbo},
  journal={arXiv preprint arXiv:2605.06481},
  year={2026}
}

@article{zhou2026tau_0,
  title={$\tau_0$-WM: A Unified Video-Action World Model for Robotic Manipulation},
  author={Zhou, Pengfei and Chen, Shengcong and Chen, Di and Wang, Jiaxu and Jin, Rongjun and Zhu, Bingwen and Pan, Yike and Gu, Songen and Wang, Kuanning and Nan, Shufeng and others},
  journal={arXiv preprint arXiv:2606.01027},
  year={2026}
}

@article{zhu2026dswam,
  title={DSWAM: A Dual-System World Action Foundation Model for Fine-Grained Robot Manipulation},
  author={Zhu, Jian and Zhang, Jianjun and Su, Taiyi and Liu, Tianbin and Wang, Zhangyuan and Xie, Kai and Huang, Zitai and Ma, Chong and He, Youzhang and Wang, Tianjian and others},
  journal={arXiv preprint arXiv:2607.04927},
  year={2026}
}

@article{pan2026selfwam,
  title={SelfWAM: A Self-Grounded Unified World Action Model for Fast Robot Control},
  author={Pan, Bikang and Liu, Fan and Lu, Haotao and Wang, Jingya and Shi, Ye},
  journal={arXiv preprint arXiv:2608.00725},
  year={2026}
}

@article{chen2026abot,
  title={Abot-m0. 5: Unified mobility-and-manipulation world action model},
  author={Chen, Ronghan and Yang, Yandan and Tang, Zuojin and Huo, Dongjie and Lin, Tong and Wu, Haoning and Liu, Haoyun and Chen, Yuzhi and Zheng, Lulu and Yuan, Botai and others},
  journal={arXiv preprint arXiv:2607.00678},
  year={2026}
}

@article{xue2026hitac,
  title={HiTac-WAM: A Hierarchical Tactile World Action Model for Contact-Rich Robot Manipulation},
  author={Xue, Chao and Zhang, Chaofan and Ma, Wenxuan and Yao, Guocai and Cui, Shaowei and Wang, Shuo},
  journal={arXiv preprint arXiv:2608.19574},
  year={2026}
}

@article{wang2026ea,
  title={EA-WM: Event-Aware World Models with Task-Specification Grounding for Long-Horizon Manipulation},
  author={Wang, Kailin and Jie, Haoxiang and Yan, Yaoyuan and Zhou, Jiacheng and Heng, Zhiyou},
  journal={arXiv preprint arXiv:2606.13053},
  year={2026}
}

@article{lin2025storm,
  title={STORM: Search-Guided Generative World Models for Robotic Manipulation},
  author={Lin, Wenjun and Zhang, Jensen and Cai, Kaitong and Wang, Keze},
  journal={arXiv preprint arXiv:2512.18477},
  year={2025}
}

@inproceedings{liang2026exopredicator,
  title={Exopredicator: Learning abstract models of dynamic worlds for robot planning},
  author={Liang, Yichao and Nguyen, Dat and Yang, Cambridge and Li, Tianyang and Tenenbaum, Joshua B and Rasmussen, Carl Edward and Weller, Adrian and Tavares, Zenna and Silver, Tom and Ellis, Kevin},
  booktitle={International Conference on Learning Representations},
  volume={2026},
  pages={49174--49214},
  year={2026}
}

@article{jiang2026flowmo,
  title={FlowMo-WM: A World Model with Object Momentum and Hidden Ambient Drift},
  author={Jiang, Yitao and Zhao, Luyang and Chen, Muhao and Balkcom, Devin},
  journal={arXiv preprint arXiv:2606.13817},
  year={2026}
}

@article{gu2026geniworld,
  title={GeniWorld: A Generalizable Interactive World Model for Robotic Manipulation via Visual Actions},
  author={Gu, Chenghao and Yu, Hanyang and Zhang, Jingbo and Lin, Haitao and Zhang, Wenyao and Wang, Jinghe and Jin, Hanglei and Xie, Shuzhao and Jiang, Jingyan and Wang, Zhi},
  journal={arXiv preprint arXiv:2608.06332},
  year={2026}
}

@article{wang2026world,
  title={World Models for Robotic Manipulation: A Survey},
  author={Wang, Fangyuan and Wang, Ziyuan and Pei, Guorui and Zhang, Mengshi and Liang, Canxi and Hu, Jun and Li, Zhongxuan and Wu, Jinsong and Han, Ning and Zhang, Zeqing and others},
  journal={arXiv preprint arXiv:2606.00113},
  year={2026}
}

@article{shao2024deepseekmath,
  title={Deepseekmath: Pushing the limits of mathematical reasoning in open language models},
  author={Shao, Zhihong and Wang, Peiyi and Zhu, Qihao and Xu, Runxin and Song, Junxiao and Bi, Xiao and Zhang, Haowei and Zhang, Mingchuan and Li, YK and Wu, Yang and others},
  journal={arXiv preprint arXiv:2402.03300},
  year={2024}
}

@book{peters2017elements,
  title={Elements of causal inference: foundations and learning algorithms},
  author={Peters, Jonas and Janzing, Dominik and Sch{\"o}lkopf, Bernhard},
  year={2017},
  publisher={The MIT press}
}

@book{hernan2010causal,
  title={Causal inference},
  author={Hern{\'a}n, Miguel A and Robins, James M},
  year={2010},
  publisher={CRC Boca Raton, FL}
}

@article{liu2026flow,
  title={Flow-grpo: Training flow matching models via online rl},
  author={Liu, Jie and Liu, Gongye and Liang, Jiajun and Li, Yangguang and Liu, Jiaheng and Wang, Xintao and Wan, Pengfei and Zhang, Di and Ouyang, Wanli},
  journal={Advances in neural information processing systems},
  volume={38},
  pages={40783--40818},
  year={2026}
}

@inproceedings{zhao2025embodied,
  title={Embodied-r: Collaborative framework for activating embodied spatial reasoning in foundation models via reinforcement learning},
  author={Zhao, Baining and Wang, Ziyou and Fang, Jianjie and Gao, Chen and Man, Fanhang and Cui, Jinqiang and Wang, Xin and Chen, Xinlei and Li, Yong and Zhu, Wenwu},
  booktitle={Proceedings of the 33rd ACM International Conference on Multimedia},
  pages={11071--11080},
  year={2025}
}

@article{cao2026z,
  title={Z-1: Efficient Reinforcement Learning for Vision-Language-Action Models},
  author={Cao, Lang and Chen, Renhong and Li, Luyi and Wang, Peng and Peng, Mofan and Li, Yitong},
  journal={arXiv preprint arXiv:2606.31846},
  year={2026}
}

@inproceedings{lee2021causal,
  title={Causal reasoning in simulation for structure and transfer learning of robot manipulation policies},
  author={Lee, Tabitha E and Zhao, Jialiang Alan and Sawhney, Amrita S and Girdhar, Siddharth and Kroemer, Oliver},
  booktitle={2021 IEEE International Conference on Robotics and Automation (ICRA)},
  pages={4776--4782},
  year={2021},
  organization={IEEE}
}

@inproceedings{ameperosa2025rocoda,
  title={Rocoda: Counterfactual data augmentation for data-efficient robot learning from demonstrations},
  author={Ameperosa, Ezra and Collins, Jeremy A and Jain, Mrinal and Garg, Animesh},
  booktitle={2025 IEEE International Conference on Robotics and Automation (ICRA)},
  pages={13250--13256},
  year={2025},
  organization={IEEE}
}

@article{zhang2026cofactvla,
  title={CofactVLA: Deconfounding Vision-Language-Action Models via Counterfactual Intervention},
  author={Zhang, Yan and Wu, Yinan and Duan, Haoran and Han, Jungong},
  journal={arXiv preprint arXiv:2608.04396},
  year={2026}
}

@article{peng2026fact,
  title={FACT: Failure-Aware Causal Training for World-Action Models},
  author={Peng, Quanquan and Liang, Yutong and Yan, Rui and Hansen, Nicklas and Wang, Xiaolong},
  journal={arXiv preprint arXiv:2608.10232},
  year={2026}
}

@article{nasiriany2024robocasa,
  title={Robocasa: Large-scale simulation of everyday tasks for generalist robots},
  author={Nasiriany, Soroush and Maddukuri, Abhiram and Zhang, Lance and Parikh, Adeet and Lo, Aaron and Joshi, Abhishek and Mandlekar, Ajay and Zhu, Yuke},
  journal={arXiv preprint arXiv:2406.02523},
  year={2024}
}

@article{liu2023libero,
  title={Libero: Benchmarking knowledge transfer for lifelong robot learning},
  author={Liu, Bo and Zhu, Yifeng and Gao, Chongkai and Feng, Yihao and Liu, Qiang and Zhu, Yuke and Stone, Peter},
  journal={Advances in Neural Information Processing Systems},
  volume={36},
  pages={44776--44791},
  year={2023}
}

@article{ali2025world,
  title={World simulation with video foundation models for physical ai},
  author={Ali, Arslan and Bai, Junjie and Bala, Maciej and Balaji, Yogesh and Blakeman, Aaron and Cai, Tiffany and Cao, Jiaxin and Cao, Tianshi and Cha, Elizabeth and Chao, Yu-Wei and others},
  journal={arXiv preprint arXiv:2511.00062},
  year={2025}
}

@article{kim2026cosmos,
  title={Cosmos policy: Fine-tuning video models for visuomotor control and planning},
  author={Kim, Moo Jin and Gao, Yihuai and Lin, Tsung-Yi and Lin, Yen-Chen and Ge, Yunhao and Lam, Grace and Liang, Percy and Song, Shuran and Liu, Ming-Yu and Finn, Chelsea and others},
  journal={arXiv preprint arXiv:2601.16163},
  year={2026}
}

@article{ke20243d,
  title={3d diffuser actor: Policy diffusion with 3d scene representations},
  author={Ke, Tsung-Wei and Gkanatsios, Nikolaos and Fragkiadaki, Katerina},
  journal={arXiv preprint arXiv:2402.10885},
  year={2024}
}

@article{ze20243d,
  title={3d diffusion policy: Generalizable visuomotor policy learning via simple 3d representations},
  author={Ze, Yanjie and Zhang, Gu and Zhang, Kangning and Hu, Chenyuan and Wang, Muhan and Xu, Huazhe},
  journal={arXiv preprint arXiv:2403.03954},
  year={2024}
}

@article{chi2025diffusion,
  title={Diffusion policy: Visuomotor policy learning via action diffusion},
  author={Chi, Cheng and Xu, Zhenjia and Feng, Siyuan and Cousineau, Eric and Du, Yilun and Burchfiel, Benjamin and Tedrake, Russ and Song, Shuran},
  journal={The International Journal of Robotics Research},
  volume={44},
  number={10-11},
  pages={1684--1704},
  year={2025},
  publisher={Sage Publications Sage UK: London, England}
}

@article{donat2025towards,
  title={Towards fusing point cloud and visual representations for imitation learning},
  author={Donat, Atalay and Jia, Xiaogang and Huang, Xi and Taranovic, Aleksandar and Blessing, Denis and Li, Ge and Zhou, Hongyi and Zhang, Hanyi and Lioutikov, Rudolf and Neumann, Gerhard},
  journal={arXiv preprint arXiv:2502.12320},
  year={2025}
}

@misc{li2025unifiedvideoactionmodel,
      title={Unified Video Action Model}, 
      author={Shuang Li and Yihuai Gao and Dorsa Sadigh and Shuran Song},
      year={2025},
      eprint={2503.00200},
      archivePrefix={arXiv},
      primaryClass={cs.RO},
      url={https://arxiv.org/abs/2503.00200}, 
}

@article{liang2025video,
  title={Video generators are robot policies},
  author={Liang, Junbang and Tokmakov, Pavel and Liu, Ruoshi and Sudhakar, Sruthi and Shah, Paarth and Ambrus, Rares and Vondrick, Carl},
  journal={arXiv preprint arXiv:2508.00795},
  year={2025}
}

@article{bjorck2025gr00t,
  title={Gr00t n1: An open foundation model for generalist humanoid robots},
  author={Bjorck, Johan and Casta{\~n}eda, Fernando and Cherniadev, Nikita and Da, Xingye and Ding, Runyu and Fan, Linxi and Fang, Yu and Fox, Dieter and Hu, Fengyuan and Huang, Spencer and others},
  journal={arXiv preprint arXiv:2503.14734},
  year={2025}
}

@article{team2024octo,
  title={Octo: An open-source generalist robot policy},
  author={Team, Octo Model and Ghosh, Dibya and Walke, Homer and Pertsch, Karl and Black, Kevin and Mees, Oier and Dasari, Sudeep and Hejna, Joey and Kreiman, Tobias and Xu, Charles and others},
  journal={arXiv preprint arXiv:2405.12213},
  year={2024}
}

@article{kim2024openvla,
  title={Openvla: An open-source vision-language-action model},
  author={Kim, Moo Jin and Pertsch, Karl and Karamcheti, Siddharth and Xiao, Ted and Balakrishna, Ashwin and Nair, Suraj and Rafailov, Rafael and Foster, Ethan and Lam, Grace and Sanketi, Pannag and others},
  journal={arXiv preprint arXiv:2406.09246},
  year={2024}
}

@inproceedings{zheng2025tracevla,
  title={Tracevla: Visual trace prompting enhances spatial-temporal awareness for generalist robotic policies},
  author={Zheng, Ruijie and Liang, Yongyuan and Huang, Shuaiyi and Gao, Jianfeng and Daum{\'e} III, Hal and Kolobov, Andrey and Huang, Furong and Yang, Jianwei},
  booktitle={International Conference on Learning Representations},
  volume={2025},
  pages={54277--54296},
  year={2025}
}

@article{qu2025spatialvla,
  title={Spatialvla: Exploring spatial representations for visual-language-action model},
  author={Qu, Delin and Song, Haoming and Chen, Qizhi and Yao, Yuanqi and Ye, Xinyi and Ding, Yan and Wang, Zhigang and Gu, JiaYuan and Zhao, Bin and Wang, Dong and others},
  journal={arXiv preprint arXiv:2501.15830},
  year={2025}
}

@article{bu2025univla,
  title={Univla: Learning to act anywhere with task-centric latent actions},
  author={Bu, Qingwen and Yang, Yanting and Cai, Jisong and Gao, Shenyuan and Ren, Guanghui and Yao, Maoqing and Luo, Ping and Li, Hongyang},
  journal={arXiv preprint arXiv:2505.06111},
  year={2025}
}

@article{li2026causal,
  title={Causal world modeling for robot control},
  author={Li, Lin and Zhang, Qihang and Luo, Yiming and Yang, Shuai and Wang, Ruilin and Han, Fei and Yu, Mingrui and Gao, Zelin and Xue, Nan and Zhu, Xing and others},
  journal={arXiv preprint arXiv:2601.21998},
  year={2026}
}

@book{pearl2009causality,
  title={Causality},
  author={Pearl, Judea},
  year={2009},
  publisher={Cambridge university press}
}
